\documentclass[letterpaper, 10 pt, conference]{lib/ieeeconf}
\IEEEoverridecommandlockouts

\usepackage{cite}
\usepackage{amsmath,amssymb,amsfonts}
\usepackage{algorithmic}
\usepackage{graphicx}
\usepackage{textcomp}
\usepackage{xcolor}
\usepackage{booktabs}       % professional-quality tables
\usepackage{dblfloatfix}    % To enable figures at the bottom of page

\usepackage{mathtools, tabulary, multirow, hhline, multicol}
\usepackage{filecontents,lipsum}

\usepackage{times}
\usepackage{epsfig}
\usepackage[pagebackref=true,breaklinks=false,letterpaper=true,colorlinks,bookmarks=false]{hyperref}

\begin{document}

%%%%%%%%% TITLE
\title{RADNet: A Deep Neural Network Model for Robust Perception in Moving Autonomous Systems}

\author{Burhan A. Mudassar, Sho Ko, Maojingjing Li, Priyabrata Saha, Saibal Mukhopadhyay\\
% Georgia Institute of Technology\\
% 266 Ferst Drive NW, Atlanta GA 30332, USA\\
\thanks{{All authors are affiliated with the School of ECE, Georgia Institute of Technology, Atlanta, GA, USA. Email: \tt\small \{burhan.mudassar, mli399, sko45, priyabratasaha\} @ gatech.edu}, {\tt\small saibal@ece.gatech.edu}
}}

\maketitle
%\thispagestyle{empty}

%%%%%%%%% ABSTRACT

\begin{abstract}
Interactive autonomous applications require robustness of the perception engine to artifacts in unconstrained videos. In this paper, we examine the effect of camera motion on the task of action detection. We develop a novel ranking method to rank videos based on the degree of global camera motion. For the high ranking camera videos we show that the accuracy of action detection is decreased. We propose an action detection pipeline that is robust to the camera motion effect and verify it empirically. Specifically, we do actor feature alignment across frames and couple global scene features with local actor-specific features. We do feature alignment using a novel formulation of the Spatio-temporal Sampling Network (STSN) but with multi-scale offset prediction and refinement using a pyramid structure. We also propose a novel input dependent weighted averaging strategy for fusing local and global features. We show the applicability of our network on our dataset of moving camera videos with high camera motion (MOVE dataset) with a 4.1\% increase in frame mAP and 17\% increase in video mAP. 

\end{abstract}

%%%%%%%%% BODY TEXT

\section{Introduction}

% An important component of video understanding is the task of spatio-temporal action localization in which actions are classified and localized both spatially and temporally. It is 

Interactive robotic and autonomous applications require detecting objects as well as the actions being performed by them \cite{jiang2014modeling, rezazadegan2017action}. For example, an autonomous car not only needs to detect pedestrians but the action being performed by them to determine its own course of action. Action detection is an important component of the perception engine and it has received considerable progress in recent years due in part to deep-learning based architectures. The effective solutions repurpose deep learning based object detection architectures such as Faster RCNN and SSD and add temporal information to them \cite{peng2016multi, kalogeiton2017action, he2018generic, simonyan2014two, singh2017online, saha2017amtnet, singh2018tramnet, zhao2019dance}. Yet, despite such advances, spatio-temporal action detection in unconstrained environments remains a hard problem due to video-specific artifacts such as motion blurring and jitter. Particularly for moving platforms, a common cause for the above artifacts is due to relative motion between the actor and the camera. Previously, camera motion has only been explored in the context of the task of action classification \cite{wu2011action, jain2013better, wang2013action, avgerinakis2015moving} and not action detection. In this work, we take a step towards understanding the role of camera motion in action detection and computational steps to achieve robustness to it.

% The temporal information is added either by processing stacks of frames (tubelets) and/or adding optical flow information to the architecture (two-stream architecture \cite{gkioxari2015finding}). 

\begin{table*}[b]
    \centering
    \resizebox{1.0\linewidth}{!}{
    \begin{tabular}{|l|c|c|c|}
        \hline
         Method & Global Features & Feature Alignment & End Task\\
         \hline
         CoupleNet \cite{zhu2017couplenet} & Concat & - & Frame Object Detection\\
         \hline
         FGFA \cite{zhu2017flow} & - & Optical Flow Warping & Video Object Detection\\
         \hline
         STSN \cite{bertasius2018object} & - & Deformable Convolutions Single Scale & Video Object Detection\\
         \hline
         TramNET \cite{singh2018tramnet} & - & Proposal linking & Video Action Detection\\
         \hline
         RTPR \cite{li2018recurrent} & Concat & - & Video Action Detection\\
         \hline
         \textbf{This Work} & Weighted Avg. & Deformable Convolutions. Refinement using Pyramids & Video Action Detection\\
         \hline 
    \end{tabular}}
    \caption{Comparison with Prior Work}
    \label{tab:algo-improvements}
\end{table*}

We focus on the task of spatio-temporal action detection in videos with high amounts of camera motion. To quantify the degree of camera motion we propose a novel ranking method that uses dense optical flow. Next, we observe that current action detection models fail in challenging videos with high amounts of camera motion. We hypothesize the reduction in accuracy is due to reduced amounts of overlap of features of the actor across frames and the absence of discriminative actor features. Inspired by camera motion compensated features for action classification \cite{wang2013action, jain2013better, avgerinakis2015moving}, we propose the addition of a fully learnable feature alignment module that predicts spatiotemporal offsets and uses deformable convolutions for compensation \cite{bertasius2018object}. The concept of deformable convolutions for feature alignment is motivated by the STSN which predicts offsets at a single scale but unlike the STSN we predict offsets at multiple scales and propose a novel offset refinement step by creating a pyramid of offsets. In addition to that, we propose the addition of contextual information \cite{zhu2017couplenet, li2018recurrent} but with a novel input-dependent weighted averaging strategy for fusion of local and global features. 
We propose RADNet, a clip-based action detection architecture that takes in a stacked set of frames $f_t...f_{t+k}$ and generates feature maps at various scales (specifically 6 here) for all the frames within the clip. The feature maps at each scale and at each time step are aligned with respect to a reference frame within the clip using our multi-scale deformable convolution and refinement procedure. The aligned features are then fused with global scene features which are then used for prediction.

% The offsets at multiple scales are refined using a novel refinement procedure that constructs a pyramid of offsets through upsampling and addition. 
Once trained, our model can handle arbitrary camera motions, an idea that we verify by training on existing action detection datasets (UCF101-24 \cite{soomro2012ucf101}) and evaluating on videos with high camera motion. We create a custom dataset of real-world moving camera videos with spatio-temporal annotation of actions. These videos comprise our \textit{MOVE dataset}, which we will release. These videos contain a diverse array of actions in challenging scenarios with high degrees of camera motion. We verify the camera motion both qualitatively and quantitatively using our ranking method. Empirically, we observe that our model improves frame mAP and video mAP by 4.1\% and 17\% respectively on the MOVE dataset. The experiment validates that once trained on any action detection dataset, our model can be generalized to consider arbitrary camera motion.

In summary, our contributions are 
\begin{itemize}
    \itemsep0em 
    \item We construct a novel ranking method for quantifying degree of camera motion within a video and characterize the effect of camera motion on the accuracy of action detection.
    \item We propose a feature alignment module that predicts offsets at multiple scales and refines them by creating a pyramid of offsets.
    \item We propose a novel global feature fusion strategy that uses weighted averaging of local and global features.
    \item We propose a combined model with feature alignment and global feature fusion that improves overall action detection accuracy.
    \item We collect a dataset of challenging internet videos (MOVE dataset) with high degree of camera motions and annotate the actions within them and show the efficacy of our proposed network on it. 
\end{itemize}

%------------------------------------------------------------------------
% \pagebreak

\section{Related Work} \label{sec:related}

\textbf{Action Detection} The effective approach in literature is to adapt object detection architectures such as RCNN and SSD and incorporate temporal information to the architecture \cite{gkioxari2015finding, weinzaepfel2015learning, peng2016multi, suman16bmvc, feichtenhofer2016convolutional, kalogeiton2017action, saha2017amtnet, singh2018tramnet, singh2018predicting, he2018generic, zhao2019dance}. The temporal information is added either by adding optical flow as an additional modality to create a two-stream framework \cite{gkioxari2015finding} or by processing a stack of frames \cite{kalogeiton2017action, saha2017amtnet, singh2018tramnet} or by using 3D CNNs as feature extractors \cite{hou2017end, gu2018ava, zhang2019structured}.

In fully supervised approaches, ROAD \cite{singh2017online} is a single frame approach using SSD networks trained with annotated actions. ACT \cite{kalogeiton2017action} and TRAMNet \cite{singh2018tramnet} are SSD-based architectures while AMTNet \cite{saha2017amtnet} and TPN \cite{he2018generic} are Faster R-CNN-based architectures that use multiple frames. In addition, multi-task formulations \cite{singh2018predicting, song2019tacnet}, recurrent classifiers \cite{li2018recurrent} and early/hybrid fusion of optical flow and RGB \cite{zhao2019dance, yang2019step} are used to improve the action detection performance. The tubelet or frame level detections from these models are linked together to form action tubes through various methods including EM/Viterbi \cite{suman16bmvc, singh2017online, kalogeiton2017action} tracking by detection \cite{weinzaepfel2015learning} or learning-based \cite{yang2019step}. This paper uses the ACT model \cite{kalogeiton2017action} for benchmarking on action detection in moving camera sequences as it is a simple and powerful baseline.

\textbf{Actions in Moving Camera Sequences} The role of camera motion has been studied for action classification \cite{avgerinakis2015moving, wu2011action, wang2013action}. These approaches separate out camera-induced and object-induced features such as superpixels \cite{avgerinakis2015moving}, dense trajectories 
\cite{wang2013action} through homography \cite{avgerinakis2015moving, wang2013action} or low-rank optimization \cite{wu2011action}. Other approaches include using human detectors \cite{rezazadegan2017action} or a CRF \cite{tokmakov2017learning} to crop object boundaries and use the cropped features to perform classification. \textbf{However, Action detection in moving camera sequences has not received much attention in prior work}.

\begin{figure*}[t]
    \centering
    \includegraphics[width=1.0\linewidth]{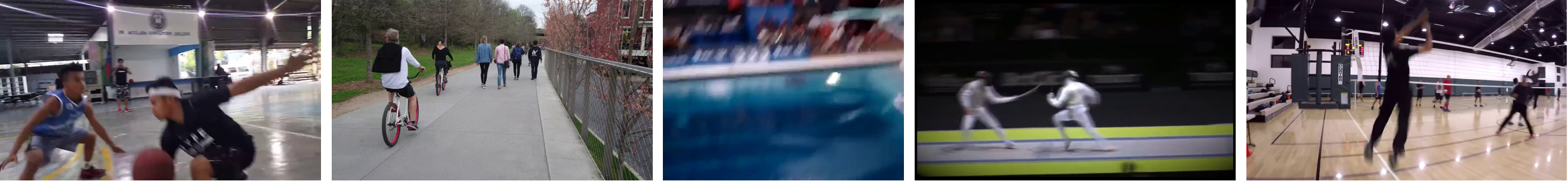}
    \caption{Some sample frames from the videos in the MOVE dataset. The videos contain diverse actions but with strong camera motion and jitter that leads to video artifacts such as motion blurring.}
    \label{fig:move_videos}
\end{figure*}

\begin{figure*}[t]
    \centering
    \includegraphics[width=1.0\linewidth]{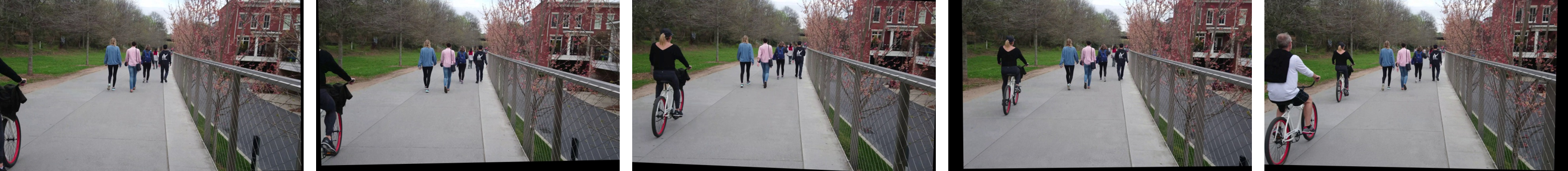}
    \caption{A video after image stabilization. The movement of the borders shows the level of compensation applied to stabilize the video}
    \label{fig:move_video_stablized}
\end{figure*}

\textbf{Datasets} Action detection datasets contain both spatial and temporal labelling of actions. JHMDB-21 \cite{cheron2015p} contains 928 clips of 21 everyday actions such as \textit{catch, pick, walk, and wave}. UCFSports \cite{soomro2012ucf101} contains 150 videos of 10 sports-based actions such as \textit{diving, running, kicking}. UCF101-24 \cite{soomro2012ucf101} is a challenging dataset of 3207 untrimmed videos with 24 actions. The actions in UCF101-24 are also sports-centric i.e. \textit{diving, cricket, diving, biking}, etc. AVA \cite{gu2018ava} is a large-scale dataset of over 80 actions in 430 15-minute video clips. \textbf{The videos in these datasets are either captured from a static viewpoint or have very smooth camera motions}. 
\vspace{-0.5em}
\subsection{Our contributions in Context of Prior Work} 

 A summary of the qualitative comparisons with prior work is given in Table \ref{tab:algo-improvements}.

\textbf{Feature Alignment in Video Data:} FGFA \cite{zhu2017flow} aligns features from multiple frames to a keyframe by using optical flow based warping. In STSN \cite{bertasius2018object} the optical flow warping step is replaced with deformable convolutions \cite{dai2017deformable}. DT\&T \cite{feichtenhofer2017detect} track the detected objects and use cross-correlated feature maps for detection. Non-local Neural Networks \cite{wang2018non} do non-local mean filtering over space and time for video action classification. Lei et al. use deformable temporal operations in which they predict temporal offsets for selecting feature maps for prediction \cite{lei2018temporal} but they do not predict spatial offsets. 

Our automated feature alignment step is most similar to the STSN \cite{bertasius2018object} proposed for video object detection. The key difference is that we generate offsets at multiple scales. Additionally, we refine the offsets by upsampling the coarse offsets and adding to the finer-scale offsets similar to a pyramid. The pyramid approach is similar to FPN \cite{lin2017feature} but we create a pyramid of offsets and not features. An orthogonal approach is used by TramNET \cite{singh2018tramnet} that aligns the final predicted boxes. Unlike their approach we do the alignment step at the feature map level.

\textbf{Global Features:} Average fusion of scene and local features is used in CoupleNet \cite{zhu2017couplenet} and RTPR \cite{li2018recurrent} does concatenation of the local and global feature maps. We develop a weighted averaging scheme for fusion.

\section{Action Detection in Moving Camera Sequences}

We focus on the problem of action detection for moving camera sequences. First, we develop a camera motion detection model based on dense optical flow. The videos containing actions are then ranked with the camera motion model. A baseline action detector is used to benchmark the action detection metrics.

\subsection{Camera Motion Detection Model}

\begin{figure}[t]
    \centering
    \includegraphics[width=1.0\linewidth]{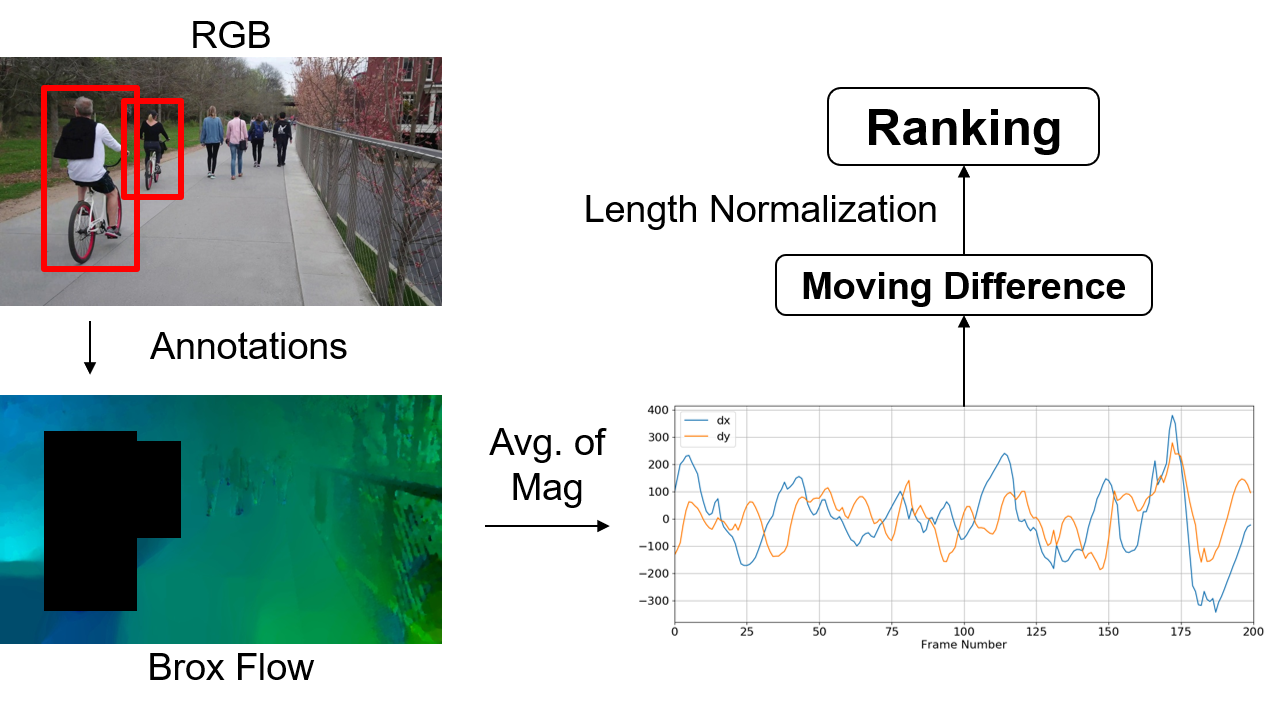}
    \caption{Model for ranking videos by camera motion content}
    \label{fig:ranking-procedure}
\end{figure}

Our model relies on dense optical flow to model the global camera motion between frames (Figure \ref{fig:ranking-procedure}). The dense optical flow is computed using the Brox Method \cite{brox2004high}. The average magnitude of the optical flow is stored for each frame (since we do not care about the direction of the apparent motion). Next, regions containing human motions are masked out as they are not always in agreement with the camera motion and may even provide erroneous estimates \cite{wang2013action}. The masking is performed using the bounding box annotations. A moving difference of the average flow is taken and normalized by the length of the video to generate the ranking (Equations \ref{eq:flow} and \ref{eq:ranking}). Videos with a higher ranking are considered to have a higher degree of camera motion. We verify the ranking through visual inspection. 
\vspace{-0.5em}
\begin{equation} \label{eq:flow}
\small    flow(t) = \frac{1}{x*y}*\sum_{x} \sum_{y} ||I_{masked}(t,x,y)||
\end{equation}
\vspace{-0.5em}
\begin{equation} \label{eq:ranking}
\small    rank = \frac{1}{nframes}*\sum_{i=1}^{nframes-1} |flow(i+1) - flow(i)|
\end{equation}

\begin{figure}[t]
    \centering
    \includegraphics[width=1.0\linewidth]{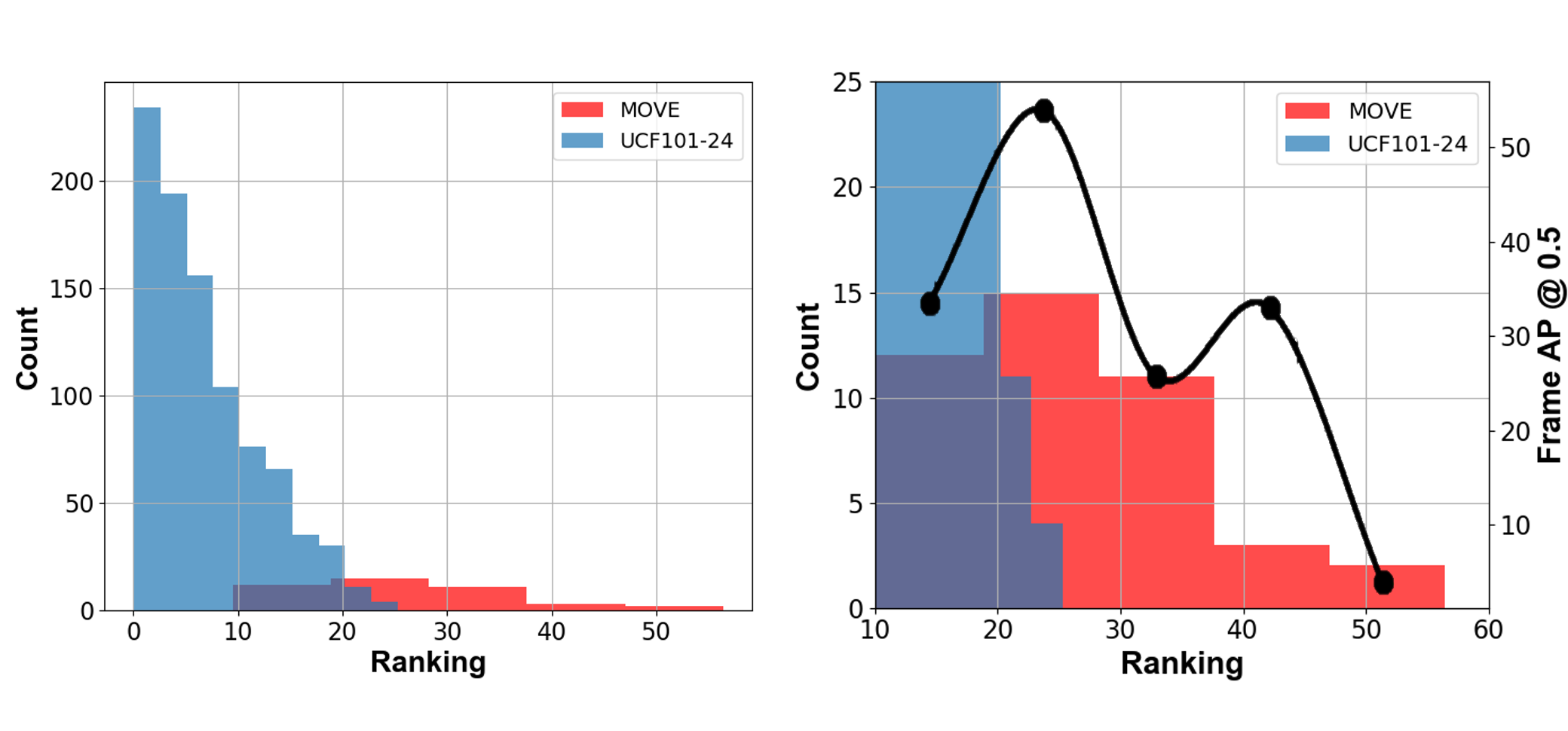}
    \caption{Left: Histogram of videos ranked according to the degree of  camera motion content. Right: Zoomed area shows count of videos with larger camera motion content and the black line shows frame AP at a threshold of 0.5 for the different clusters on the MOVE dataset with the baseline ACT ResNet50.}
    \label{fig:move_rankings}
    \vspace{-1em}
\end{figure}

We compared our ranking with a video stabilization algorithm that uses Shi-Tomasi Corner Features and Pyramid Lucas-Kanade flow \cite{vidstab2019}. The stabilization transforms are summed up and length normalized to generate the video ranking. Our ranking method gave a higher ranking to videos with jitter and irregular camera motions. The top ranking videos in the UCF-101-24 dataset with both rankings are shown in the supplementary material.

\begin{figure*}[t]    
    \centering
    \includegraphics[width=0.95\linewidth]{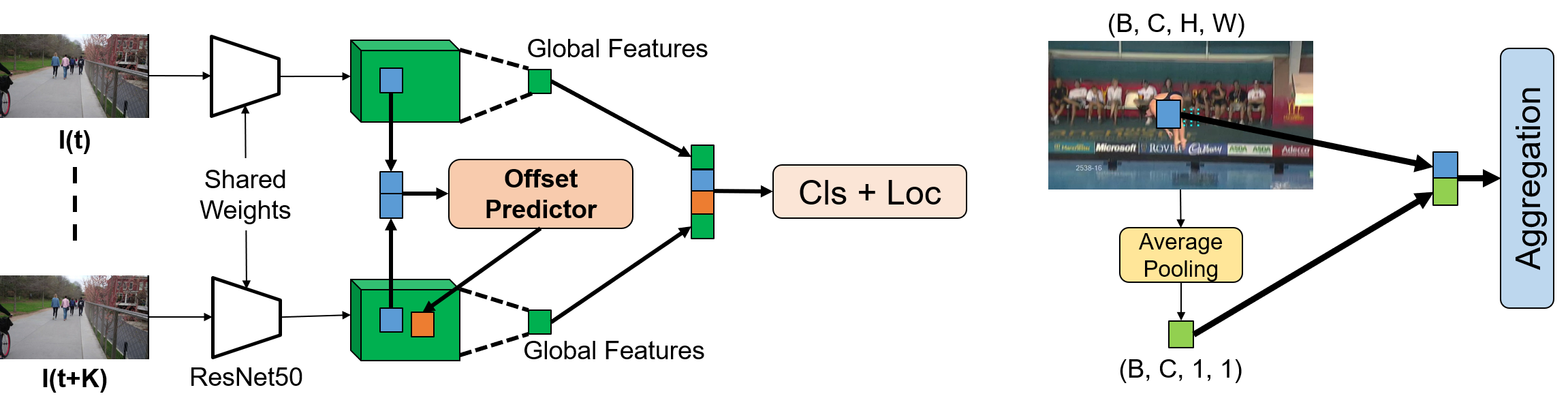}
    \caption{The RADNet action detection model tailored for detection in moving camera sequences. Feature alignment is done by predicting likely locations of the objects in different frames using an offset predictor and the global scene features are aggregated with the local actor features.}
     \label{fig:full-model}
\end{figure*}

\subsection{MOVE: A Dataset of Moving Camera Sequences}

The camera motion ranking model shows that only a small subset of videos in the UCF101-24 dataset contains videos with high degree of camera motion. This motivates us to collect videos containing actions and with natural camera motions. We collected a dataset of moving camera videos from Youtube to empirically evaluate performance of action detection models on real-world sequences with natural camera motion. We chose videos containing activities that overlapped with the UCF-101-24 dataset and we collected 43 sequences with various actions and high degrees of camera motion. Some of these sequences are shown in Figure \ref{fig:move_videos}. A sample video is also shown with video stabilization applied to it in Figure \ref{fig:move_video_stablized} to show the degree of camera motion present. In these videos, actions were annotated spatially and temporally. Of the 24 actions present in UCF-24, we were able to collect videos containing 15 of those actions with clip lengths varying from 2 to 12 seconds. A comparison of the rankings for videos in UCF101-24 and the MOVE dataset in Figure \ref{fig:move_rankings} shows the stark difference in estimated motion for the videos in UCF101-24 and our dataset. Over 23 videos in our dataset rank higher than the highest ranked videos in UCF101-24. \textbf{Our objective is not to train on this data but to simply use for evaluation.}

\section{The RADNET Model}

The RADNet model for action detection builds upon the ACT \cite{kalogeiton2017action} framework for action detection. The network uses a stacked set of frames (clip/tubelet) to generate predictions. For each frame within the clip, feature maps (at multiple scales for one-stage detection) are generated by applying a convolutional backbone to the input frames and the generated feature maps (at the same scale) are combined together in the time dimension. Anchor cuboids are generated apriori for each location in the feature map by extending anchor boxes in the same spatial location over time. For each anchor cuboid, box regression and action classification is performed. The key difference is the addition of two layers. An automated feature alignment layer with multi-scale refinement spatially aligns features from multiple frames while a global feature fusion layer incorporates contextual information with local-actor specific features.

\subsection{Feature Alignment with Multi-Scale Refinement}

The feature alignment step aligns features spatially across the time dimension. Formally, given a stacked set of features $X_{stacked}$ corresponding to a stacked set of frames through timesteps $N-K$ to $N$, a prediction module $M$ makes class predictions and localizations for every point $(i,j)$ on that feature map (Equation \ref{eq: prediction}) for time step $N$. 

% $M$ can be fully convolutional or it can be a RNN-type module. 
\vspace{-0.5em}
\begin{equation}
\small
    X_{stacked} = DepthConcat(X_{N-K}, X_{N-K+1}, ..., X_N)
\end{equation}
\vspace{-0.5em}
\begin{equation} \label{eq: prediction}
\small    cls, loc (N,i,j) = M(X_{stacked})
\end{equation}

For a point on the feature map $(i,j)$ our objective is to transform $X_{N-K}$ through $X_N$ so that features relevant to $(i,j)$ are passed through (Equation \ref{eq: transform}). The transformation is achieved by tuning the receptive field of the filters at that location by adding an offset from the set of predicted offsets $\Delta P$. 
\vspace{-0.5em}
\begin{equation} \label{eq: transform}
\small    X(t,i,j) = \sum X(t, p + \Delta p(t,i,j) ),  {t \in T}
\end{equation}

\begin{figure}[t]
    \centering
    \includegraphics[width=1.0\linewidth]{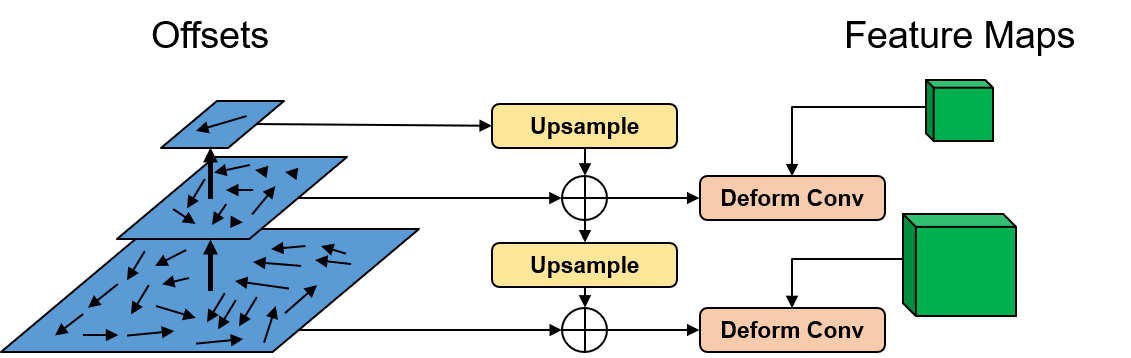}
    \caption{Refinement of predicted offsets at different scales}
    \label{fig:coarse2finearch}
\end{figure}

The set of offsets $\Delta P$ is computed from the stacked set of feature maps $X_{stacked}$ for every point in the sampling grid $G$. Let $F$ be the function that predicts offsets for each time step for every spatial point in the feature map. We apply $F$ to get $\Delta P$ as shown in Equation \ref{eq:per-frame-offset}. 

\begin{equation} \label{eq:per-frame-offset}
\small    \Delta P = F(X_{stacked})
\end{equation}

$F$ can be any transformation such as optical flow between frames. For a fully end-to-end learning framework, $F$ needs to be differentiable. The STSN framework using deformable convolutions is adopted to model $F$. Briefly, a 2D convolution layer predicts offsets which are then used to offset the sampling locations for the convolution operator. However, convolution is a very local operation with limited receptive fields i.e. 3x3 and cannot model the global camera motion effectively. The solution is to create a pyramid representation of the offsets. The formulation for the pyramid is trivial \cite{bouguet2001pyramidal} and is adapted to RADNet as follows. 

\begin{figure}[t]
    \centering
    \includegraphics[width=1.0\linewidth]{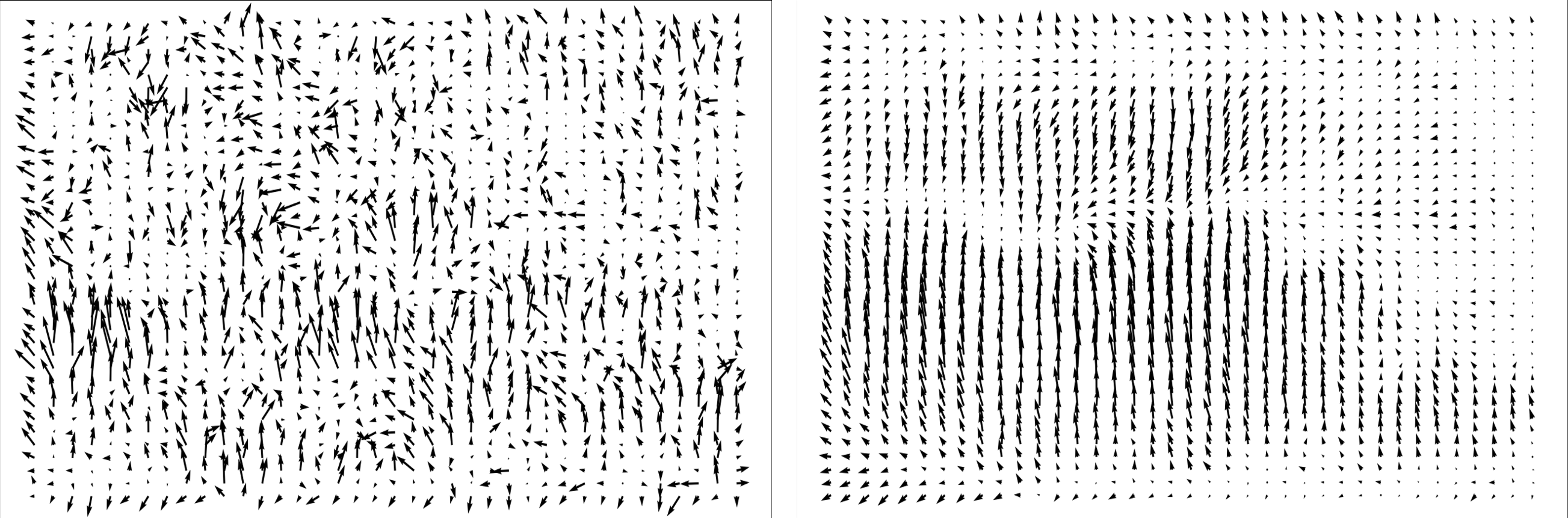}
    \caption{Before and after refinement of offsets at the finest scale}
    \label{fig:coarse2fineresults}
    \vspace{-1em}
\end{figure}

Let $O$ be the set of offsets with $o^{k}$ being the offsets computed at the $kth$ scale. In SSD, the feature maps at 6 scales are used for prediction thus there are 6 levels of offsets in $O$ with scales ranging from 38 x 38 to 1 x 1. The offsets at the coarsest scale (1 x 1) model the average camera motion component. They are then upsampled and added to the finer scales recursively. The up-sampled is performed using bilinear sampling. The block diagram for the refinement is shown in Figure \ref{fig:coarse2finearch}. A qualitative comparison before and after refinement is shown in Figure \ref{fig:coarse2fineresults}. The noisy local motion components are smoothed using the refinement procedure.

\subsection{Global and Local Feature Fusion}

\begin{figure}[t]
    \centering
    \includegraphics[width=0.9\linewidth]{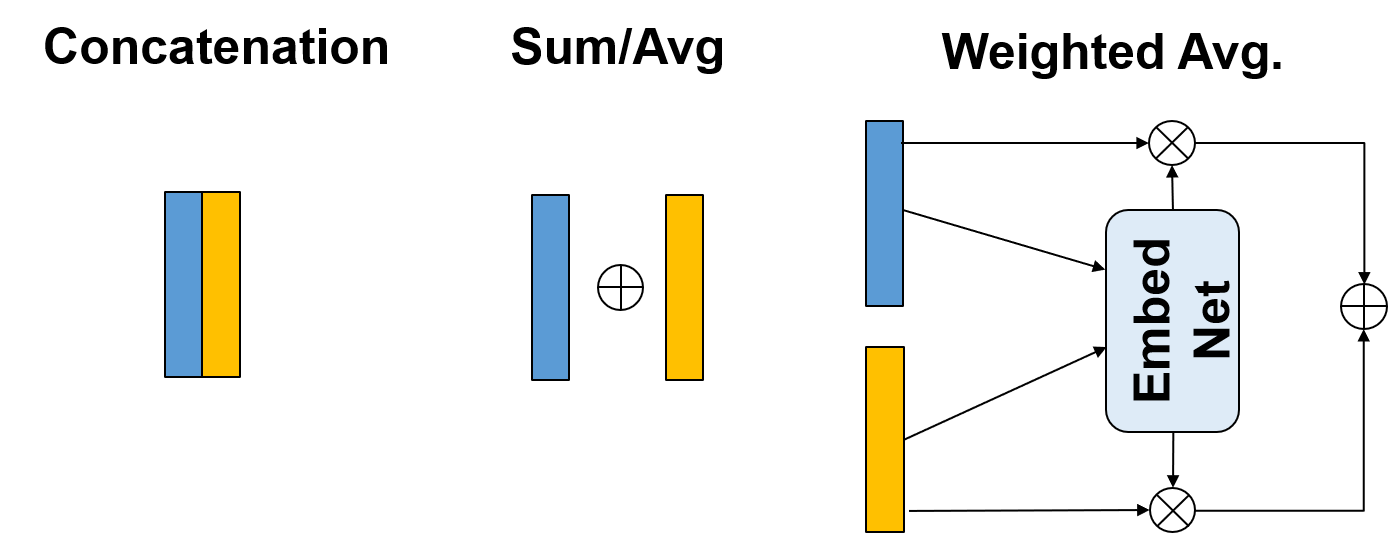}
    \caption{Aggregation strategies for local features and global features}
    \label{fig:arch-globalfusion}
\end{figure}

To incorporate contextual information, we extract features corresponding to the whole scene and fuse them with the local actor specific features $f_{local}$. The global features $f_{global}$ are generated by average pooling the feature map at that scale. The global feature map is then aggregated with the local feature maps for prediction. Multiple fusion strategies for aggregation are implemented including concatenation, averaging and weighted averaging as shown in Figure \ref{fig:arch-globalfusion}.

\begin{figure*}[t]
    \centering
    \includegraphics[width=\linewidth]{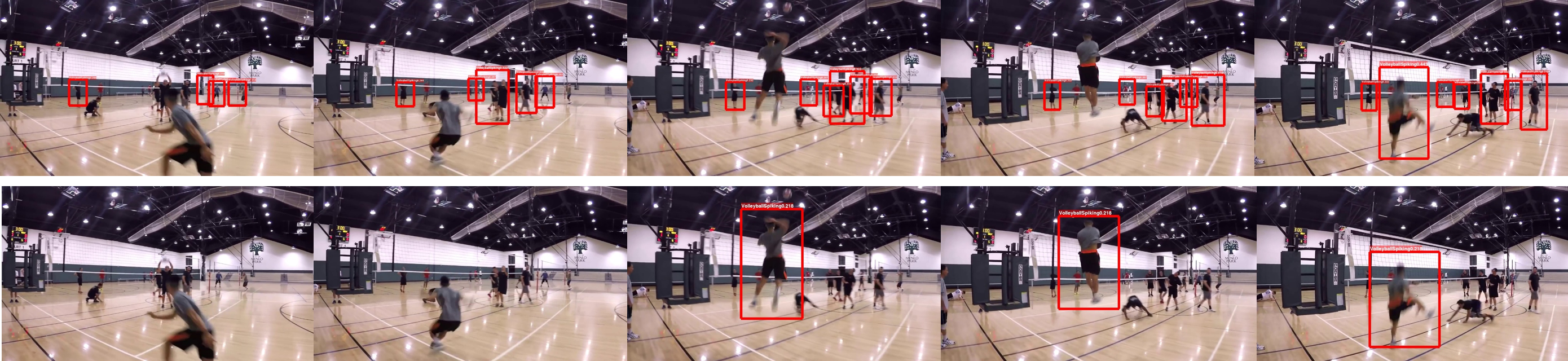}
    \caption{Results on the MOVE dataset after post-processing of detections. Only the action labelled for the video is shown for visual clarity. Top Row: With the ACT detector. Bottom: Our model.
    In the volleyball video, the ACT detector does not detect the correct actor and has many false positives}
    \label{fig:results-move}
\end{figure*}

\textbf{1. Concatenation} In this scheme, the local and global feature maps are concatenated along the channel dimension. The width of the next layer needs to be doubled to accommodate this.

\textbf{2. Average/Summation} The local and global feature maps are averaged or summed up along the channel dimension. The width of the next layer does not change in this scheme.

\textbf{3. Weighted Averaging} A novel weighted averaging scheme is implemented that dynamically weighs the global and local features for every input sample and for every input spatial location. The weighted averaging is implemented by using an embedding function to generate weights for the actor features and global features. Formally let the embedding function be $f_{embed}$ and let the actor features weight be $l^i$ and the global feature be $g^i$ for the $ith$ input sample. The embeddings for both are generated by applying $f_{embed}$ as given in Equation \ref{eq:embed1}.

\begin{equation} \label{eq:embed1}
\small    l_{embed}^i, g_{embed}^i = f_{embed}(l^i), f_{embed}(g^i)
\end{equation}

The normalized weight embeddings $w_{normalized}^i$ are then generated by applying the softmax function over concatenation of $l_{embed}$ and $g_{embed}$ (Equation \ref{eq:embed2}). 

\begin{equation} \label{eq:embed2}
\small    w_{normalized}^i = softmax(concat(l_{embed}^i, g_{embed}^i))
\end{equation}

The $w_{normalized}$ is then used to do the weighted summation of the global and local features as given in Equation \ref{eq:embed4} to get the output feature map $p$ for prediction.

\begin{equation} \label{eq:embed4}
\small    p^i = l^i * w^i_{normalized}[0] + g^i * w^i_{normalized}[1]
\end{equation}
% \vspace{-0.5em}

We use a 3 layer DNN \cite{zhu2017flow} EmbedNet to implement the $f_{embed}$ function. The EmbedNet takes in the local features and the global features of dimension $(B,C,H,W)$. The output of EmbedNet is a one-channel feature map $(B,1,H,W)$. The local and global embeddings are then concatenated and the softmax is applied to get a normalized weighted feature map. The global and local weights are then multiplied with the global and local features respectively.

\section{Experimental Results and Analysis} \label{sec:experiments}

We evaluate the different configurations of RADNet on the MOVE Dataset. In addition, we also benchmark RADNet on existing action detection datasets JHMDB-21 and UCF101-24. For accuracy benchmarking, frame AP and video AP metrics are used. Frame AP evaluates correctness of detections only at the frame level while video AP checks for temporal consistency of detections for the duration of the video.

\textbf{Implementation Details :} We use the ResNet-50 backbone in all of our models. 3 extra residual blocks are added and feature maps from res3, res4, res5 and the extra layers are used. All images are resized to 300 x 300 for training and testing. The model is trained with a step learning schedule and a starting learning rate of 1e-3 for a total of 130000 iterations and a batch size of 8. The feature alignment layers are added to the output of all scales.

\begin{table}[!t]%[!htbp]%[t]
    \centering
    \resizebox{1.0\linewidth}{!}{
    \begin{tabular}{|c|l||c|c||c|c|}
        \hline
        \multirow{2}{*}{detector} & \multirow{2}{*}{method} & \multicolumn{2}{|c||}{JHMDB-21} &  \multicolumn{2}{|c|}{UCF-101-24}  \\ 
        \cline{3-6} 
        &  & 0.2 & 0.5:0.95 & 0.2 & 0.5:0.95\\ \hline \hline
        
        % \multirow{3}{*}{Faster R-CNN}  & ~\cite{suman16bmvc}  & 72.6 & 40.0 &  66.7 & 14.4\\ 
        %   & Generic~\cite{he2018generic}  & 79.7 & - &  71.6 & - \\
        %   & RTPR~\cite{li2018recurrent} & \textbf{82.7} & - & \textbf{77.9} & -\\ 
        %   & \textcolor{red}{\textbf{ours}} & 72.8 & \textbf{43.6} & 75.1 & \textbf{19.3} \\
        %  \cline{1-6}
        \multirow{3}{*}{SSD}
         & ROAD ~\cite{singh2017online}  & 73.8 &  41.6 &  73.5 & 20.4 \\  
         & ACT~\cite{kalogeiton2017action}  & 74.2 & 44.8 &  
         76.5 & 23.4 \\
         & TramNet $\Delta$1~\cite{singh2018tramnet} & - & - & 79.0 & 23.9 \\
         & AMTNet-L~\cite{singh2018tramnet} & - & - & \textbf{79.4} & 23.4 \\
         & TACNET \cite{song2019tacnet} & 74.1 & 44.8 & 77.5 & 24.1 \\
         & \textbf{ours} & \textbf{75.1} & \textbf{48.3} & 78.4 & \textbf{24.5} \\
        \hline
    \end{tabular}}
    \caption{Comparison Results on video mAP of our method to the state of the art at various detection thresholds. 
    }
    \label{table:sotavideoAP}
    \vspace{-4mm}
\end{table}

\begin{table}[tbh] 
    \centering
    \resizebox{0.95\linewidth}{!}{
    \begin{tabular}{l|c|c|c|c}
        \multirow{2}{*}{Network} & frameAP & \multicolumn{3}{c}{Video mAP} \\ % & \multirow{2}{*}{Params} & \multirow{2}{*}{FLOPS} \\ 
        \cline{2-5}
        & 0.5 & 0.1 & 0.2 & 0.3 \\
        \hline
        ACT ResNet-50 & 27.7 & 57.8 & 52.3 & 48.7\\% & 35.5 & 17.5  \\
        \hline
        \multicolumn{5}{c}{\textbf{Feature Alignment}} \\
        \hline
        +STSN\cite{bertasius2018object} & 27.1  & 55.9 & 50.5 & 43.9\\%  & 93.1 & 39.1 \\
        +STSN+refinement & 31.1 & 67.2  & 63.9  & 57.7\\%  & 93.1 & 39.1\\
    \end{tabular}}
    \caption{Comparison on the MOVE dataset with and without feature alignment  \label{table:move-align}}
\end{table}

\begin{table}[tbh] 
    \centering
    \resizebox{0.95\linewidth}{!}{
    \begin{tabular}{l|c|c|c|c}
        \multirow{2}{*}{Network} & frameAP & \multicolumn{3}{c}{Video mAP} \\ % & \multirow{2}{*}{Params} & \multirow{2}{*}{FLOPS} \\ 
        \cline{2-5}
        & 0.5 & 0.1 & 0.2 & 0.3 \\
        \hline
        ACT ResNet-50 & 27.7 & 57.8 & 52.3 & 48.7\\% & 35.5 & 17.5  \\
        \hline
        \multicolumn{5}{c}{\textbf{Global Feature Fusion}}\\
        \hline
        Concat & 29.0 & 54.6 & 48.6 & 46.7\\ %& 43.4 & 19.4 \\
        Avg & 29.4 & 59.9 & 56.4 & 45.0\\ % & 35.5 & 17.5\\
        Weighted Avg & 29.2 & 66.1 & 60.5 & 51.9\\ % & 63.6 & 28.9 \\
    \end{tabular}}
    \caption{Comparison on the MOVE dataset with and without global feature fusion  \label{table:move-global}}
\end{table}

\begin{table}[tbh] 
    \centering
    \resizebox{0.95\linewidth}{!}{
        \begin{tabular}{ccc|c|c|c}
        STSN\cite{bertasius2018object} & global & refinement & frame AP & Params & FLOPS \\ 
        \hline
        & & & 27.7 & 35.5 & 17.5\\
        \checkmark & & & 27.1 & 93.1 & 39.1\\
        % \hline
        \checkmark & & \checkmark & 31.1 & 93.1 & 39.1\\
        & \checkmark & & 29.2 & 63.6 & 28.9\\
        \checkmark & \checkmark & & 31.6 & 121.4 & 49.6\\
        \checkmark & \checkmark & \checkmark & \textbf{31.8} & 121.4 & 49.6\\
        \hline
    \end{tabular}}
    \caption{Ablation study on the MOVE Dataset  \label{table:move-ablation}}
\end{table}

\begin{table}[tbh] 
    \centering
    \resizebox{0.95\linewidth}{!}{
    \begin{tabular}{l|c|c|c|c}
        \multirow{2}{*}{Network} & frameAP & \multicolumn{3}{c}{Video mAP} \\ % & \multirow{2}{*}{Params} & \multirow{2}{*}{FLOPS} \\ 
        \cline{2-5}
        & 0.5 & 0.1 & 0.2 & 0.3 \\
        \hline
        ACT ResNet-50 & 27.7 & 57.8 & 52.3 & 48.7\\% & 35.5 & 17.5  \\
        RADNet & 31.8 & 74.9 & 70.8 & 65.7\\
    \end{tabular}}
    \caption{Comparison on the MOVE Dataset with RADNeT and prior work  \label{table:move-prior}}
\end{table}

\subsection{Evaluation on Existing Datasets}

Evaluation of video mAP on the UCF101-24 and JHMDB-21 datasets (Table \ref{table:sotavideoAP}) is performed. For direct comparison, both RGB and optical flow modalities are used and we use the two-stream network formulation \cite{gkioxari2015finding}. Similar to prior work, we average the results for all splits for JHMDB and for UCF101 we report the results for the first split only. For the JHMDB dataset, we improve the video mAP at all thresholds. For UCF101-24, we improve the video mAP at at 0.5:0.95 by 0.4\%. The small increase in video mAP is attributed to the fact that the datasets contain few moving camera videos so our network does not improve the performance by much.

\subsection{Evaluation on MOVE Dataset}

Evaluation on the MOVE dataset is only performed on the RGB images and with the RGB-trained network.

\textbf{Feature Alignemnt} The effect of feature alignment is discussed here (Table \ref{table:move-align}). In the first experiment the STSN module without refinement is added to the network. The accuracy metrics are decreased. We attribute this to the inaccurate offsets being computed. With the addition of the refinement the frame AP is boosted by 3.4\% and the video AP is also improved by 9.0\% at a threshold of 0.3 This validates quantitatively the advantage of the refinement step. It also comes at a minimal cost with no addition in the parameters of the model.

\textbf{Global Fusion Strategies:} Next, the different global fusion strategies are evaluated in the network (Table \ref{table:move-global}). The weighted averaging shows a clear advantage over the concatenation and simple averaging with increase in video mAP (66.1\% vs. 57.8\%) at a threshold of 0.1 and 51.9\% vs 48.7\% at a threshold of 0.3.

\textbf{Ablation Studies:} Ablation studies were performed on the MOVE dataset with different configurations of the model (Table \ref{table:move-ablation}). It is interesting to note that the combined STSN and global feature fusion model improve the frame AP more than their individual contributions. Thus, their effects are complementary. The weighted averaging scheme is used for fusion in all the configurations. The highest frame AP and video AP are observed with the offset refinement and weighted averaging global feature fusion (31.8\%). For RADNet we use this configuration to compare with ACT and improve frame AP by 4.1\% and video AP by 17\% at a threshold of 0.3 (Table \ref{table:move-prior}).

\section{Conclusion} 
In this paper, we have presented the problem of spatio-temporal action detection in moving camera sequences. We have collected and presented a test dataset containing moving camera sequences with annotated actions. We have shown a novel action detection architecture that is robust to the effect of camera motion in action detection. For future work, we intend to collect more videos for testing and possibly for training as well. In addition we will investigate the two-stream model for fusing optical flow input in the moving camera context. We hope that this will encourage further research into this domain.

\section*{Acknowledgements}

The research reported here was supported in part by the Defense
Advanced Research Projects Agency (DARPA) under
contract number HR0011-17-2-0045. The views and conclusions
contained herein are those of the authors and should not
be interpreted as necessarily representing the official policies or endorsements, either expressed or implied, of DARPA. 

{\small
\bibliographystyle{lib/ieee}
\bibliography{Bibliography.bib}

\begin{thebibliography}{10}\itemsep=-1pt

\bibitem{vidstab2019}
A python package to stabilize videos using opencv.
\newblock \url{https://github.com/AdamSpannbauer/python_video_stab}.
\newblock Accessed: 2019-11-15.

\bibitem{avgerinakis2015moving}
K.~Avgerinakis, K.~Adam, A.~Briassouli, and Y.~Kompatsiaris.
\newblock Moving camera human activity localization and recognition with
  motionplanes and multiple homographies.
\newblock In {\em Image Processing (ICIP), 2015 IEEE International Conference
  on}, pages 2085--2089. IEEE, 2015.

\bibitem{bertasius2018object}
G.~Bertasius, L.~Torresani, and J.~Shi.
\newblock Object detection in video with spatiotemporal sampling networks.
\newblock In {\em Proceedings of the European Conference on Computer Vision
  (ECCV)}, pages 331--346, 2018.

\bibitem{bouguet2001pyramidal}
J.-Y. Bouguet et~al.
\newblock Pyramidal implementation of the affine lucas kanade feature tracker
  description of the algorithm.
\newblock {\em Intel Corporation}, 5(1-10):4, 2001.

\bibitem{brox2004high}
T.~Brox, A.~Bruhn, N.~Papenberg, and J.~Weickert.
\newblock High accuracy optical flow estimation based on a theory for warping.
\newblock In {\em European conference on computer vision}, pages 25--36.
  Springer, 2004.

\bibitem{cheron2015p}
G.~Ch{\'e}ron, I.~Laptev, and C.~Schmid.
\newblock P-cnn: Pose-based cnn features for action recognition.
\newblock In {\em Proceedings of the IEEE international conference on computer
  vision}, pages 3218--3226, 2015.

\bibitem{dai2017deformable}
J.~Dai, H.~Qi, Y.~Xiong, Y.~Li, G.~Zhang, H.~Hu, and Y.~Wei.
\newblock Deformable convolutional networks.
\newblock In {\em Proceedings of the IEEE International Conference on Computer
  Vision}, pages 764--773, 2017.

\bibitem{feichtenhofer2016convolutional}
C.~Feichtenhofer, A.~Pinz, and A.~Zisserman.
\newblock Convolutional two-stream network fusion for video action recognition.
\newblock In {\em Proceedings of the IEEE Conference on Computer Vision and
  Pattern Recognition}, pages 1933--1941, 2016.

\bibitem{feichtenhofer2017detect}
C.~Feichtenhofer, A.~Pinz, and A.~Zisserman.
\newblock Detect to track and track to detect.
\newblock In {\em Proceedings of the IEEE International Conference on Computer
  Vision}, pages 3038--3046, 2017.

\bibitem{gkioxari2015finding}
G.~Gkioxari and J.~Malik.
\newblock Finding action tubes.
\newblock In {\em Proceedings of the IEEE conference on computer vision and
  pattern recognition}, pages 759--768, 2015.

\bibitem{gu2018ava}
C.~Gu, C.~Sun, D.~A. Ross, C.~Vondrick, C.~Pantofaru, Y.~Li,
  S.~Vijayanarasimhan, G.~Toderici, S.~Ricco, R.~Sukthankar, et~al.
\newblock Ava: A video dataset of spatio-temporally localized atomic visual
  actions.
\newblock In {\em Proceedings of the IEEE Conference on Computer Vision and
  Pattern Recognition}, pages 6047--6056, 2018.

\bibitem{he2018generic}
J.~He, Z.~Deng, M.~S. Ibrahim, and G.~Mori.
\newblock Generic tubelet proposals for action localization.
\newblock In {\em 2018 IEEE Winter Conference on Applications of Computer
  Vision (WACV)}, pages 343--351. IEEE, 2018.

\bibitem{hou2017end}
R.~Hou, C.~Chen, and M.~Shah.
\newblock An end-to-end 3d convolutional neural network for action detection
  and segmentation in videos.
\newblock {\em arXiv preprint arXiv:1712.01111}, 2017.

\bibitem{jain2013better}
M.~Jain, H.~Jegou, and P.~Bouthemy.
\newblock Better exploiting motion for better action recognition.
\newblock In {\em Proceedings of the IEEE conference on computer vision and
  pattern recognition}, pages 2555--2562, 2013.

\bibitem{jiang2014modeling}
Y.~Jiang and A.~Saxena.
\newblock Modeling high-dimensional humans for activity anticipation using
  gaussian process latent crfs.
\newblock In {\em Robotics: Science and systems}, pages 1--8, 2014.

\bibitem{kalogeiton2017action}
V.~Kalogeiton, P.~Weinzaepfel, V.~Ferrari, and C.~Schmid.
\newblock Action tubelet detector for spatio-temporal action localization.
\newblock {\em ICCV}, 2017.

\bibitem{lei2018temporal}
P.~Lei and S.~Todorovic.
\newblock Temporal deformable residual networks for action segmentation in
  videos.
\newblock In {\em Proceedings of the IEEE Conference on Computer Vision and
  Pattern Recognition}, pages 6742--6751, 2018.

\bibitem{li2018recurrent}
D.~Li, Z.~Qiu, Q.~Dai, T.~Yao, and T.~Mei.
\newblock Recurrent tubelet proposal and recognition networks for action
  detection.
\newblock In {\em Proceedings of the European conference on computer vision
  (ECCV)}, pages 303--318, 2018.

\bibitem{lin2017feature}
T.-Y. Lin, P.~Doll{\'a}r, R.~Girshick, K.~He, B.~Hariharan, and S.~Belongie.
\newblock Feature pyramid networks for object detection.
\newblock In {\em Proceedings of the IEEE conference on computer vision and
  pattern recognition}, pages 2117--2125, 2017.

\bibitem{peng2016multi}
X.~Peng and C.~Schmid.
\newblock Multi-region two-stream r-cnn for action detection.
\newblock In {\em European Conference on Computer Vision}, pages 744--759.
  Springer, 2016.

\bibitem{rezazadegan2017action}
F.~Rezazadegan, S.~Shirazi, B.~Upcrofit, and M.~Milford.
\newblock Action recognition: From static datasets to moving robots.
\newblock In {\em Robotics and Automation (ICRA), 2017 IEEE International
  Conference on}, pages 3185--3191. IEEE, 2017.

\bibitem{saha2017amtnet}
S.~Saha, G.~Singh, and F.~Cuzzolin.
\newblock Amtnet: Action-micro-tube regression by end-to-end trainable deep
  architecture.
\newblock In {\em Proceedings of the IEEE International Conference on Computer
  Vision}, pages 4414--4423, 2017.

\bibitem{suman16bmvc}
S.~Saha, G.~Singh, M.~Sapienza, P.~H. Torr, and F.~Cuzzolin.
\newblock Deep learning for detecting multiple space-time action tubes in
  videos.
\newblock In {\em Proceedings of the British Machine Vision Conference}, 2016.

\bibitem{simonyan2014two}
K.~Simonyan and A.~Zisserman.
\newblock Two-stream convolutional networks for action recognition in videos.
\newblock In {\em Advances in neural information processing systems}, pages
  568--576, 2014.

\bibitem{singh2018predicting}
G.~Singh, S.~Saha, and F.~Cuzzolin.
\newblock Predicting action tubes.
\newblock In {\em Proceedings of the European Conference on Computer Vision
  (ECCV)}, pages 0--0, 2018.

\bibitem{singh2018tramnet}
G.~Singh, S.~Saha, and F.~Cuzzolin.
\newblock Tramnet-transition matrix network for efficient action tube
  proposals.
\newblock In {\em Asian Conference on Computer Vision}, pages 420--437.
  Springer, 2018.

\bibitem{singh2017online}
G.~Singh, S.~Saha, M.~Sapienza, P.~H. Torr, and F.~Cuzzolin.
\newblock Online real-time multiple spatiotemporal action localisation and
  prediction.
\newblock In {\em ICCV}, pages 3657--3666, 2017.

\bibitem{song2019tacnet}
L.~Song, S.~Zhang, G.~Yu, and H.~Sun.
\newblock Tacnet: Transition-aware context network for spatio-temporal action
  detection.
\newblock In {\em Proceedings of the IEEE Conference on Computer Vision and
  Pattern Recognition}, pages 11987--11995, 2019.

\bibitem{soomro2012ucf101}
K.~Soomro, A.~R. Zamir, and M.~Shah.
\newblock Ucf101: A dataset of 101 human actions classes from videos in the
  wild.
\newblock {\em arXiv preprint arXiv:1212.0402}, 2012.

\bibitem{tokmakov2017learning}
P.~Tokmakov, K.~Alahari, and C.~Schmid.
\newblock Learning motion patterns in videos.
\newblock In {\em Proceedings of the IEEE Conference on Computer Vision and
  Pattern Recognition}, pages 3386--3394, 2017.

\bibitem{wang2013action}
H.~Wang and C.~Schmid.
\newblock Action recognition with improved trajectories.
\newblock In {\em Proceedings of the IEEE international conference on computer
  vision}, pages 3551--3558, 2013.

\bibitem{wang2018non}
X.~Wang, R.~Girshick, A.~Gupta, and K.~He.
\newblock Non-local neural networks.
\newblock In {\em Proceedings of the IEEE Conference on Computer Vision and
  Pattern Recognition}, pages 7794--7803, 2018.

\bibitem{weinzaepfel2015learning}
P.~Weinzaepfel, Z.~Harchaoui, and C.~Schmid.
\newblock Learning to track for spatio-temporal action localization.
\newblock In {\em Proceedings of the IEEE international conference on computer
  vision}, pages 3164--3172, 2015.

\bibitem{wu2011action}
S.~Wu, O.~Oreifej, and M.~Shah.
\newblock Action recognition in videos acquired by a moving camera using motion
  decomposition of lagrangian particle trajectories.
\newblock In {\em 2011 International conference on computer vision}, pages
  1419--1426. IEEE, 2011.

\bibitem{yang2019step}
X.~Yang, X.~Yang, M.-Y. Liu, F.~Xiao, L.~S. Davis, and J.~Kautz.
\newblock Step: Spatio-temporal progressive learning for video action
  detection.
\newblock In {\em Proceedings of the IEEE Conference on Computer Vision and
  Pattern Recognition}, pages 264--272, 2019.

\bibitem{zhang2019structured}
Y.~Zhang, P.~Tokmakov, M.~Hebert, and C.~Schmid.
\newblock A structured model for action detection.
\newblock In {\em Proceedings of the IEEE Conference on Computer Vision and
  Pattern Recognition}, pages 9975--9984, 2019.

\bibitem{zhao2019dance}
J.~Zhao and C.~G. Snoek.
\newblock Dance with flow: Two-in-one stream action detection.
\newblock In {\em Proceedings of the IEEE Conference on Computer Vision and
  Pattern Recognition}, pages 9935--9944, 2019.

\bibitem{zhu2017flow}
X.~Zhu, Y.~Wang, J.~Dai, L.~Yuan, and Y.~Wei.
\newblock Flow-guided feature aggregation for video object detection.
\newblock In {\em Proceedings of the IEEE International Conference on Computer
  Vision}, volume~3, 2017.

\bibitem{zhu2017couplenet}
Y.~Zhu, C.~Zhao, J.~Wang, X.~Zhao, Y.~Wu, and H.~Lu.
\newblock Couplenet: Coupling global structure with local parts for object
  detection.
\newblock In {\em Proceedings of the IEEE International Conference on Computer
  Vision}, pages 4126--4134, 2017.

\end{thebibliography}
}

\end{document}